\documentclass[10pt,twocolumn,letterpaper]{article}

\usepackage{cvpr} 

\usepackage{graphicx}
\usepackage{amsmath}
\usepackage{amssymb}
\usepackage{booktabs}
\usepackage{float}
\usepackage[svgnames]{xcolor}
\definecolor{Teal}{RGB}{80,139,243}
\definecolor{MyOr}{RGB}{250,190,88}
\definecolor{MyGr}{RGB}{77,175,124}
\definecolor{MyRed}{RGB}{226,106,106}
\definecolor{MyPur}{RGB}{148,124,176}
\usepackage{xcolor, soul}
\usepackage[pagebackref=true,breaklinks=true,colorlinks,bookmarks=false]{hyperref}
\hypersetup{citecolor=[rgb]{0.3125,0.5429,0.9492}}

\usepackage[normalem]{ulem}

\newcommand\blfootnote[1]{%
  \begingroup
  \renewcommand\thefootnote{}\footnote{#1}%
  \addtocounter{footnote}{-1}%
  \endgroup
}
\usepackage[capitalize]{cleveref}
\crefname{section}{Sec.}{Secs.}
\Crefname{section}{Section}{Sections}
\Crefname{table}{Table}{Tables}
\crefname{table}{Tab.}{Tabs.}

\def\cvprPaperID{3896} 
\def\confName{CVPR}
\def\confYear{2024}

\usepackage{overpic}
\usepackage{enumitem} 
\usepackage{overpic} 
\usepackage{color}
\usepackage{colortbl}
\usepackage{array}
\usepackage{booktabs}
\usepackage{multirow}

\definecolor{turquoise}{cmyk}{0.65,0,0.1,0.3}
\definecolor{purple}{rgb}{0.65,0,0.65}
\definecolor{dark_green}{rgb}{0, 0.5, 0}
\definecolor{orange}{rgb}{0.8, 0.6, 0.2}
\definecolor{red}{rgb}{0.8, 0.2, 0.2}
\definecolor{darkred}{rgb}{0.6, 0.1, 0.05}
\definecolor{blueish}{rgb}{0.0, 0.3, .6}
\definecolor{light_gray}{rgb}{0.7, 0.7, .7}
\definecolor{pink}{rgb}{1, 0, 1}
\definecolor{greyblue}{rgb}{0.25, 0.25, 1}

\DeclareMathOperator*{\argmin}{arg\,min}

\usepackage{blindtext}

\renewcommand{\paragraph}[1]{\vspace{1em}\noindent\textbf{#1}.}
\begin{document}
\title{EMAGE: Towards Unified Holistic Co-Speech Gesture Generation via Expressive Masked Audio Gesture Modeling}

\author{Haiyang Liu\textsuperscript{1*} \quad 
Zihao Zhu\textsuperscript{2*}\quad 
Giorgio Becherini\textsuperscript{3} \quad 
Yichen Peng\textsuperscript{4} \quad 
Mingyang Su\textsuperscript{5} \quad \\
You Zhou \quad 
Xuefei Zhe \quad 
Naoya Iwamoto \quad 
Bo Zheng \quad
Michael J. Black\textsuperscript{3} 
 \\
\\
\textsuperscript{1}The University of Tokyo \quad
\textsuperscript{2}Keio University  \\
\textsuperscript{3}Max Planck Institute for Intelligent Systems \quad
\textsuperscript{4}JAIST \quad
\textsuperscript{5}Tsinghua University \quad
}


\twocolumn[{%
\renewcommand\twocolumn[1][]{#1}%
\maketitle
\begin{center}
    \centering
    \captionsetup{type=figure}
    \includegraphics[trim=0 0 0 0, clip,width=1.0\textwidth]{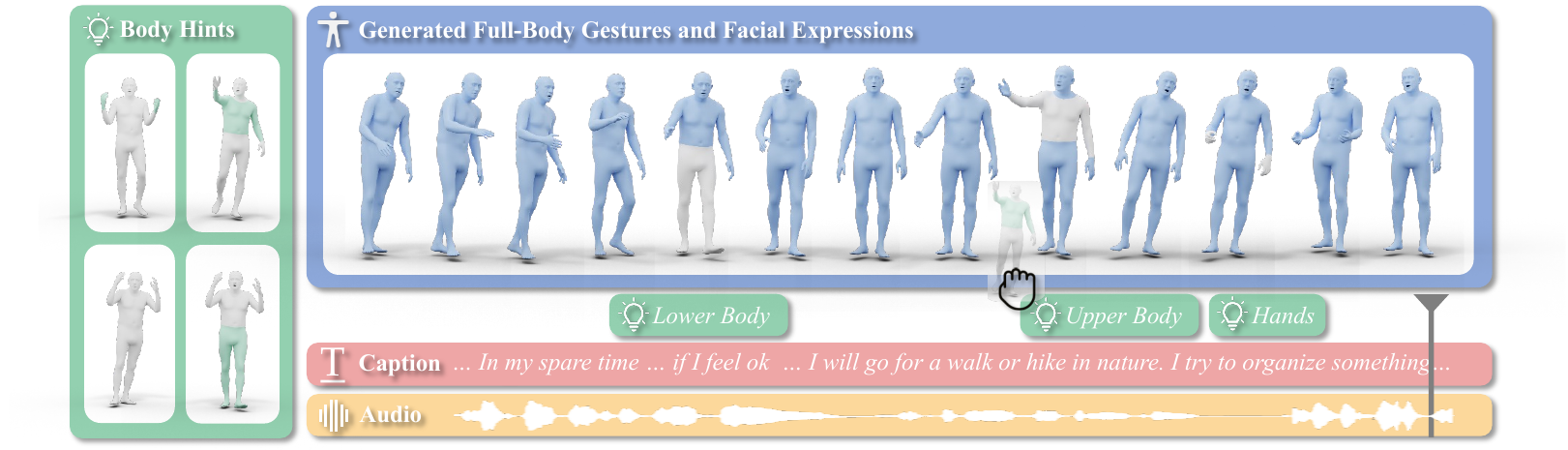}
    \vspace{-0.65cm}
    \captionof{figure}{\textbf{EMAGE.} We present a Masked Audio-Conditioned Gesture Modeling framework, along with a new holistic gesture dataset, BEAT2 (BEAT-SMPLX-FLAME), for jointly generating \textcolor{Teal}{facial expressions}, \textcolor{MyOr}{local body dynamics}, \textcolor{MyGr}{hand movements} and \textcolor{MyRed}{global translations}, conditioned on audio and a partially or completely masked \textcolor{Teal}{ge}\textcolor{MyOr}{st}\textcolor{MyGr}{ur}\textcolor{MyRed}{es}. The gray denotes visible
    gestures, and blue represents our outputs.}

    \label{fig:fig1}
\end{center}%

}]
\begin{abstract}
We propose EMAGE, a framework to generate full-body human gestures from audio and masked gestures, encompassing facial, local body, hands, and global movements. To achieve this, we first introduce BEAT2 (BEAT-SMPLX-FLAME), a new mesh-level holistic co-speech dataset. BEAT2 combines a MoShed SMPL-X body with FLAME head parameters and further refines the modeling of head, neck, and finger movements, offering a community-standardized, high-quality 3D motion captured dataset. 
EMAGE leverages masked body gesture priors during training to boost inference performance. It involves a Masked Audio Gesture Transformer, facilitating joint training on audio-to-gesture generation and masked gesture reconstruction to effectively encode audio and body gesture hints. Encoded body hints from masked gestures are then separately employed to generate facial and body movements. 
Moreover, EMAGE adaptively merges speech features from the audio's rhythm and content and utilizes four compositional VQ-VAEs to enhance the results' fidelity and diversity. Experiments demonstrate that EMAGE generates holistic gestures with state-of-the-art performance and is flexible in accepting predefined spatial-temporal gesture inputs, generating complete, audio-synchronized results. Our code and dataset are available.\blfootnote{\textsuperscript{*}Equal Contribution.}\footnote{\url{https://pantomatrix.github.io/EMAGE/}}\end{abstract}

\section{Introduction}
The path towards full-body co-speech gesture generation presents many remaining challenges.
Despite the development of various baselines, \textit{e.g.}, audio conditioned models for face \cite{fan2022faceformer,xing2023codetalker,EMOTE}, body \cite{yoon2020speech,ginosar2019learning,chhatre2023emotional}, and hand movements \cite{liu2022learning,liu2022beat,li2021audio2gestures}, along with a few attempts at merged models \cite{habibie2021learning,talkshow:yi2022generating}, the limited availability of comprehensive data and models poses an ongoing obstacle. 
To make progress in this direction, we present a new framework that can incorporate partial spatial-temporal predefined gestures and autonomously complete the remaining frames in synchronization with the audio. 
This provides a new tool for creating realistic animations of digital humans \cite{yang2023makeup,shiohara2023blendface,yang2022bareskinnet,zhou2022audio}.

To achieve this, we first require a comprehensive gesture dataset. While several datasets are available for audio-to-body \cite{yoon2020speech}, audio-to-body\&face \cite{habibie2021learning}, body-to-hands \cite{liu2022learning, liu2022beat}, and common actions \cite{AMASS:ICCV:2019,deichlerspatio}, integrating them poses challenges due to varying data formats and evaluation metrics across different sub-areas. This motivates us to establish a unified, community-standard benchmark for training and evaluating holistic gestures. 
We build upon the BEAT dataset \cite{liu2022beat}, which provides the most comprehensive mocap co-speech gesture dataset with diverse modalities. 
Despite its size and richness, the data in BEAT is not presented in a standardized format. The dataset comprises iPhone ARKit blendshape weights and Vicon skeletons, posing challenges when using the data for training; 
\textit{e.g.}, the lack of a mesh representation prevents the use of a vertex loss \cite{petrovich2021action, faceformer2022}. 
Additionally, conducting direct human perceptual evaluations on these diverse modalities is inherently complex. On the other hand, SMPL-X \cite{SMPL-X:2019} and FLAME \cite{FLAME:SiggraphAsia2017} are common mesh standards for the academic community for other human dynamics-related tasks, \textit{e.g.}, action \cite{petrovich2021action} and talking head generation \cite{VOCA2019}. Driven by the motivation to facilitate the sharing of knowledge across different tasks, we present BEAT-SMPLX-FLAME (BEAT2), which consists of two main components: \textbf{i)} Refined SMPL-X body shape and pose parameters from MoSh++ \cite{mosh:SIGASIA:2014,AMASS:ICCV:2019} with hard-coded physical priors. \textbf{ii)} Optimized high-quality FLAME head parameters. By integrating SMPL-X's body and FLAME's head, BEAT2 facilitates a comprehensive training and evaluation benchmark across multiple sub-areas of co-speech human animation generation.

\begin{table*}
\centering
\resizebox{0.99\linewidth}{!}{
\begin{tabular}{lcccccccccccccc}
              & \begin{tabular}[c]{@{}c@{}}Trinity\\Mocap\\2017 \cite{ferstl2018investigating}\end{tabular} & \begin{tabular}[c]{@{}c@{}}S2G\\S2G-2D\\2019 \cite{ginosar2019learning}\end{tabular} & \begin{tabular}[c]{@{}c@{}}Seq2Seq\\TED-2D\\2019 \cite{yoon2019robots}\end{tabular} & \begin{tabular}[c]{@{}c@{}}TWH\\Mocap\\2019 \cite{talking:lee2019talking} \end{tabular} & \begin{tabular}[c]{@{}c@{}}Trimodal \\TED-3D\\2020 \cite{yoon2020speech}\end{tabular} & \begin{tabular}[c]{@{}c@{}}VOCA\\Scan\\2020 \cite{VOCA2019}\end{tabular} & \begin{tabular}[c]{@{}c@{}}MeshTalk\\Scan\\2021 \cite{meshtalk:richard2021meshtalk}\end{tabular} & \begin{tabular}[c]{@{}c@{}}Hahibie \textit{et al.} \\S2G-3D\\2021 \cite{habibie2021learning}\end{tabular} & \begin{tabular}[c]{@{}c@{}}HA2G\\TED-3D+\\2022 \cite{ha2g:liu2022learning}\end{tabular} & \begin{tabular}[c]{@{}c@{}}BEAT\\Mocap\\2022 \cite{liu2022beat} \end{tabular} & \begin{tabular}[c]{@{}c@{}}ZEEG\\Mocap\\2022 \cite{ghorbani2022zeroeggs}\end{tabular} & \begin{tabular}[c]{@{}c@{}}Yoon. \textit{et al.}\\TED-SMPL\\2023 \cite{lu2023co}\end{tabular} & \begin{tabular}[c]{@{}c@{}}Talkshow\\S2G-SMPL\\2023 \cite{talkshow:yi2022generating}\end{tabular} & \begin{tabular}[c]{@{}c@{}}BEAT2\\Ours\\2024\end{tabular}  \\ 
\hline
Head          & -                                                            & 2D                                                        & -                                                             & -                                                        & -                                                              & 3D Scan                                                  & 3D Scan                                                      & 3D                                                                   & -                                                           & ARKit                                                     & -                                                         & -                                                                   & PGT-Mesh                                                         & MC-Mesh                                                     \\
Upper Body    & 3D                                                           & 2D                                                        & 2D                                                            & 3D                                                       & 3D                                                             & -                                                        & -                                                            & 3D                                                                   & 3D                                                          & 3D                                                        & 3D                                                        & PGT-Mesh                                                            & PGT-Mesh                                                         & MC-Mesh                                                     \\
Hands         & 3D                                                           & 2D                                                        & -                                                             & 3D                                                       & -                                                              & -                                                        & -                                                            & 3D                                                                   & 3D                                                          & 3D                                                        & 3D                                                        & PGT-Mesh                                                            & PGT-Mesh                                                         & MC-Mesh                                                     \\
Lower Body    & 3D                                                           & -                                                         & -                                                             & 3D                                                       & -                                                              & -                                                        & -                                                            & -                                                                    & -                                                           & 3D                                                        & 3D                                                        & -                                                                   & -                                                                & MC-Mesh                                                     \\
Global Motion & 3D                                                           & -                                                         & -                                                             & 3D                                                       & -                                                              & -                                                        & -                                                            & -                                                                    & -                                                           & 3D                                                        & 3D                                                        & -                                                                   & -                                                                & MC-Mesh                                                     \\
Duration (hours)     & 4                                                            & 60                                                        & 97                                                            & 20                                                       & 97                                                             & 0.5                                                      & 2                                                            & 38                                                                   & 33                                                          & 76                                                        & 4                                                         & 30                                                                  & 27                                                               & 60                                                         
\end{tabular}
}
\vspace{-0.08in}
\caption{\textbf{Comparison of Co-Speech Gesture Datasets.} We summarize gesture and face datasets in talking scenarios. `PGT' and `MC' denote Pseudo Ground Truth and Motion Capture, respectively. The BEAT2 (Ours) dataset is the largest dataset for motion-captured data, providing holistic, academic community standard mesh-level information.}
\label{tab:tab1}
\end{table*}
\begin{table*}
%
\centering
\resizebox{0.99\linewidth}{!}{
\begin{tabular}{lccccccccccccccccc}
               & \begin{tabular}[c]{@{}c@{}}S2G\\\\2019 \cite{ginosar2019learning}\end{tabular} & \begin{tabular}[c]{@{}c@{}}TriModal\\\\2020 \cite{yoon2020speech}\end{tabular} & \begin{tabular}[c]{@{}c@{}}B2H\\\\2020\cite{ng2021body2hands} \end{tabular} & \begin{tabular}[c]{@{}c@{}}TWH\\\\2021\cite{talking:lee2019talking}\end{tabular} & \begin{tabular}[c]{@{}c@{}}Hahibie\\\textit{et al.}\\2021\cite{habibie2021learning}\end{tabular} & \begin{tabular}[c]{@{}c@{}}DisCo\\\\2022\cite{liu2022disco}\end{tabular} & \begin{tabular}[c]{@{}c@{}}HA2G\\\\2022\cite{ha2g:liu2022learning}\end{tabular} & \begin{tabular}[c]{@{}c@{}}Face\\Former\\2022\cite{faceformer2022}\end{tabular} & \begin{tabular}[c]{@{}c@{}}Rhythmic\\ \\2022\cite{ao2022rhythmic}\end{tabular} & \begin{tabular}[c]{@{}c@{}}BEAT\\CaMN\\2022\cite{liu2022beat}\end{tabular} & \begin{tabular}[c]{@{}c@{}}Diff\\Gesture\\2023\cite{diffgesture:zhu2023taming}\end{tabular} & \begin{tabular}[c]{@{}c@{}}Code\\Talker\\2023\cite{xing2023codetalker}\end{tabular} & \begin{tabular}[c]{@{}c@{}}Talk\\Show\\2023\cite{talkshow:yi2022generating}\end{tabular} & \begin{tabular}[c]{@{}c@{}}DiffStlyle\\Gesture\\2023\cite{yang2023diffusestylegesture}\end{tabular} & \begin{tabular}[c]{@{}c@{}}Body\\Former\\2023\cite{pang2023bodyformer}\end{tabular} & \begin{tabular}[c]{@{}c@{}}Lively\\Speaker\\2023\cite{livelyspeaker:Zhi_2023_ICCV}\end{tabular} & \begin{tabular}[c]{@{}c@{}}EMAEG\\(Ours)\\2024\end{tabular}  \\ 
\hline
Input Gestures & -                                                   & -                                                        & Body                                                & Body                                                & -                                                             & -                                                     & -                                                    & -                                                          & -                                                            & -                                                        & -                                                           & -                                                          & -                                                        & -                                                                 & -                                                          & -                                                             & Masked                                                      \\
Head           & -                                                   & -                                                        & -                                                   & -                                                   & M                                                             & -                                                     & -                                                    & S                                                          & -                                                            & -                                                        & -                                                           & S                                                          & S                                                        & -                                                                 & -                                                          & -                                                             & C                                                           \\
Upper          & S                                                   & S                                                        & -                                                   & -                                                   & M                                                             & S                                                     & {C}                        & -                                                          & S                                                            & {C}                            & S                                                           & -                                                          & {C}                            & S                                                                 & S                                                          & S                                                             & C                                                           \\
Hands          & -                                                   & -                                                        & S                                                   & S                                                   & M                                                             & S                                                     & {C}                        & -                                                          & -                                                            & {C}                            & S                                                           & -                                                          & {C}                            & S                                                                 & -                                                          & S                                                             & C                                                           \\
Lower + Global & -                                                   & -                                                        & -                                                   & -                                                   & -                                                             & -                                                     & -                                                    & -                                                          & -                                                            & -                                                        & -                                                           & -                                                          & -                                                        & S                                                                 & -                                                          & -                                                             & C                                                          
\end{tabular}
}
\vspace{-0.08in}
\caption{\textbf{Comparison of Co-Speech Gesture Models.} We compare with previous methods for generating face or body motion trained on co-speech datasets. The first row lists their inputs, and the subsequent rows list their outputs, respectively. Different decoder designs are denoted by the initials `S' for Single, `M' for Multiple, and `C' for Cascaded. To the best of our knowledge, EMAGE (Ours) is the first to accept audio and partially or completely masked gestures, generating full-body audio-synchronized results.}

\label{tab:tab2}
\end{table*}
\begin{figure*}[]
\centerline{
\includegraphics[width=2.1\columnwidth]{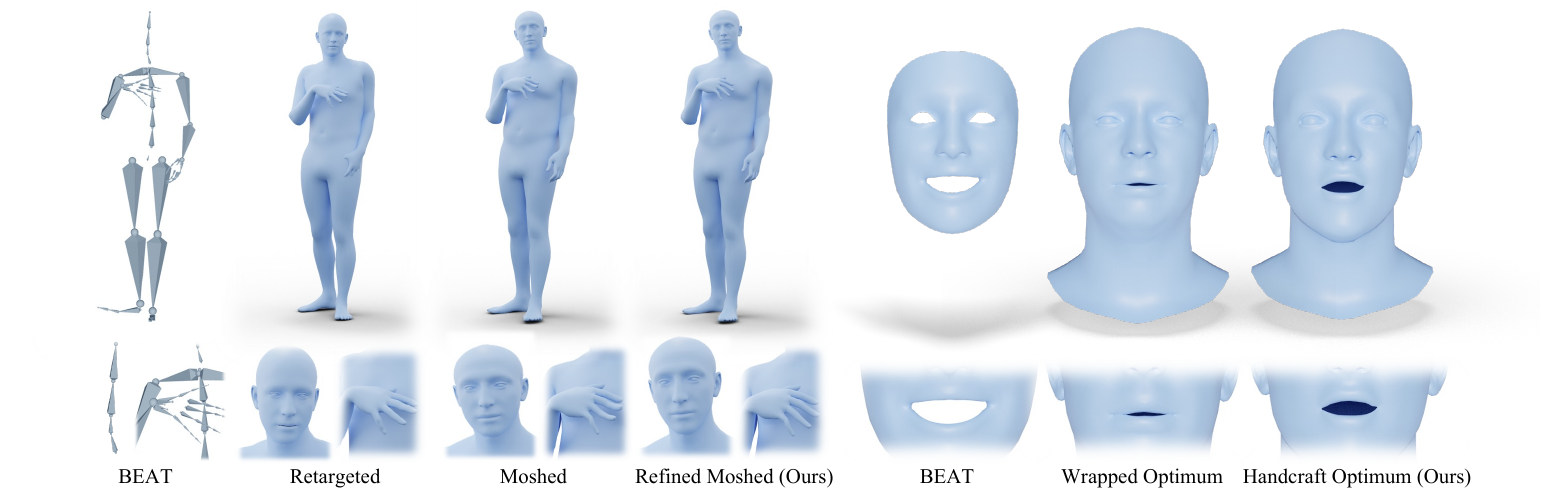}
}
\vspace{-0.1in}
\caption{
\textbf{Comparison of Data between BEAT2 and Others.} BEAT-SMPLX-FLAME presents a new mesh-level, motion-captured, holistic co-speech gesture dataset with 60h of data. \textit{Left:} We compare our refined SMPL-X body parameters  (denoted as Refined MoSh) with the original BEAT skeleton \cite{liu2022beat}, Retargeted from AutoRegPro, and initial results of Mosh++ \cite{SMPL-X:2019}. The refined results show correct neck flexion, appropriate head and neck shape ratios, and detailed finger representations. \textit{Right:} Visualization of blendshape weights from the original BEAT dataset \cite{liu2022beat} with ARKit's template, Wrapped-based, and handcrafted optimization. Our final handcrafted FLAME blendshape-based optimization demonstrates both accurate lip movement details and natural mouth shapes. 
}
\label{fig:fig2}
\vspace{-0.2cm}
\end{figure*}
With the holistic data from BEAT2, we aim to enhance the coherence of individual body parts while ensuring accurate cross-modal alignment between audio and motion; see Figure \ref{fig:fig1}. This leads to the design of Expressive Masked Audio-conditioned GEsture modeling, denoted as EMAGE, a spatial and temporal transformer-based framework. EMAGE first aggregates spatial features from masked body joints. Then, it reconstructs the latent space of pretrained gestures with a switchable gesture temporal self-attention and audio-gesture cross-attention. The selection of different forward paths enables modeling the gesture-to-gesture and audio-to-gesture prior separately and effectively. Once the reconstructed latent features are obtained, EMAGE decodes local face and body gestures from four compositional pretrained Vector Quantized-Variational AutoEncoders (VQ-VAEs) and decodes global translations from a pretrained Global Motion Predictor. The compositional VQ-VAEs are key to preserving audio-related motions during the reconstruction. Through these designs, EMAGE achieves state-of-the-art performance in generating body and face gestures. It takes as input audio and partially initialized gestures, and recoveres audio-synchronized, coherent gestures. 
Additionally, we show how EMAGE can flexibly incorporate additional, non-holistic, datasets to improve results.

Overall, our contributions are as follows:
(1) We release BEAT2, a representation-unified, mesh-level dataset that leverages MoShed SMPL-X body and FLAME head parameters.
(2) We propose EMAGE, a simple yet effective holistic gesture generation framework that generates coherent gestures with partial gesture and audio priors.
(3) EMAGE achieves state-of-the-art performance in generating both body and face gestures with only four frames of seed gestures.
(4) EMAGE showcases the use of additional non-holistic gesture datasets for training, \textit{e.g.}, Trinity and AMASS, to further improve the fidelity and diversity of the results, effectively leveraging data from different datasets.

\section{Related Work}
\noindent \textbf{Co-Speech Animation Datasets} are categorized into two types; see Table \ref{tab:tab1}: pseudo-labeled (PGT) and motion-captured (mocap). For PGT, the Speech2Gesture Dataset \cite{ginosar2019learning} utilizes OpenPose \cite{cao2019openpose} to extract 2D poses from News and Teaching videos. Subsequent works extend the dataset to 3D pose \cite{habibie2021learning} and SMPL-X \cite{talkshow:yi2022generating}. The TED Dataset \cite{yoon2019robots}, which extracts 2D pose from TED videos, has been extended to include 3D pose \cite{yoon2020speech}, fingers \cite{ha2g:liu2022learning}, and SMPL-X \cite{lu2023co}. 
Similarly, 2D, 3D landmarks, and meshes are estimated from multi-view recorded videos \cite{mead:kaisiyuan2020mead} as PGT for talking-head generation. Although pseudo-labeled approaches could theoretically extract infinite data, their accuracy is limited. For instance, the average error (in mm) for the SOTA monocular 3D pose estimation algorithm \cite{gartner2022differentiable} on the Human3.6M dataset \cite{ionescu2013human3} is 33.4, while for Vicon mocap it is 0.142 \cite{vicon}. For mocap datasets, Trinity \cite{trinity:alexanderson2020style} features one male actor with 4h of data, and TalkingWithHands \cite{talking:lee2019talking} collects conversation scenarios for two speakers. ZEEG \cite{ghorbani2022zeroeggs} considers one speaker with 12 styles. For the face, 3D scanning is accurate but limited in quantity due to cost, \textit{e.g.}, less than 3h for BIWI \cite{biwi:fanelli20103}, VOCA \cite{VOCA2019}, and MeshTalk \cite{meshtalk:richard2021meshtalk}. This leads to a balance between performance and quantity in the ARKit dataset \cite{peng2023emotalk}. The aforementioned datasets address the body and face separately. On the other hand, BEAT \cite{liu2022beat} includes both 3D pose body and ARKit \cite{arkit:dehghan2021arkitscenes} facial blendshape weights for the first time. However, BEAT lacks mesh data.

\noindent \textbf{Co-Speech Animation Models} are categorized based on their outputs; see Table \ref{tab:tab2}. In addition to the baselines and datasets \cite{ginosar2019learning, yoon2020speech, VOCA2019, meshtalk:richard2021meshtalk, talking:lee2019talking, ng2021body2hands}, several frameworks for body gestures \cite{qi2023emotiongesture,qi2023weakly, ahuja2020style,li2021audio2gestures} have improved the performance of the baselines. They are trained and evaluated by selecting specific upper body joints \cite{liu2022disco,liu2022beat,ao2022rhythmic, Ao2023GestureDiffuCLIP, ha2g:liu2022learning, livelyspeaker:Zhi_2023_ICCV, pang2023bodyformer}, or all body joints \cite{diffgesture:zhu2023taming, yang2023diffusestylegesture}. Recent improvements in facial gestures drive the vertices with transformers and discrete face priors \cite{faceformer2022, xing2023codetalker}. However, these methods only address either the face or body. Applying these methods to the full body will yield sub-optimal results as audio is differently correlated with face and body dynamics. 
Most similar to our work, Habibie \textit{et al.} \cite{habibie2021learning} employ a single audio encoder and multiple decoders for generating face and body gestures. TalkSHOW \cite{talkshow:yi2022generating} demonstrates the advantage of separating the audio encoders and decoding the body and hands using quantized codes in an auto-regressive manner. However, it lacks lower body and global motion, and there is no shared information for body and facial gesture generation. Moreover, it cannot accept partially masked body hints due to the design of a fully auto-regressive model.

\noindent \textbf{Masked Representation Learning} was first demonstrated to be effective in Natural Language Processing, with BERT-based models \cite{devlin2018bert, lan2019albert, liu2019roberta} boosting the performance of learned word embeddings in downstream tasks through a combination of masked language modeling and transformer architecture. Subsequently, Masked AutoEncoders \cite{he2022masked} expanded masked image modeling to computer vision by removing and inpainting image patches. This concept of masked representation learning has since been employed in other modalities, \textit{e.g.}, video \cite{metatrans17:dosovitskiy2020image, metatrans18:chen2022vision, metatrans19:liu2021swin}, audio \cite{metatrans6:gong2021ast,liu2020reinforcement,liu2021improving}, and point clouds \cite{metatrans9:zhao2021point, metatrans20:yu2022point, metatrans21:qian2022pix4point}. Most related to our work, MotionBERT \cite{motionbert2022} proposes a spatial-temporal transformer to learn a robust motion representation for classification-based tasks by masking 2D pose. In contrast to their method, we target robust motion features for conditional motion generation tasks, requiring a balance in training between multiple modalities, \textit{e.g.}, audio and gesture.

\section{BEAT2}
\label{sec:other}
This section introduces how we obtain unified mesh-level data, \textit{i.e.}, SMPL-X \cite{SMPL-X:2019} and FLAME \cite{FLAME:SiggraphAsia2017} parameters, from the original BEAT dataset \cite{liu2022beat}. BEAT utilizes a Vicon motion capture system \cite{vicon} and releases 78-joint skeleton-level Biovision Hierarchy (BVH) \cite{bvh} motion files. Their facial capture system uses ARKit \cite{arkit:dehghan2021arkitscenes} with a depth camera on the iPhone 12 Pro, extracting $\mathbb{R}^{51}$ blendshape weights. 
These blendshapes are designed based on the Facial Action Coding System (FACS), which is widely adopted. 
However, both the motion and facial data lack mesh-level details (see Figure \ref{fig:fig2}), \textit{e.g.}, shapes and vertices. 
\subsection{Body Parameters Initialization via MoSh++}
We initialize the SMPL-X body shape and pose parameters from BEAT mocap marker data using MoSh++ \cite{AMASS:ICCV:2019}. Given the captured markers positions $\mathbf{m} \in \mathbb{R}^{T \times K \times 3}$, predefined markers position offsets $\mathbf{d} \in \mathbb{R}^{K \times 3}$, and user-defined vertices-to-markers function $\mathcal{H}$, we aim to obtain body shape $\beta \in \mathbb{R}^{300}$, poses $\theta \in \mathbb{R}^{T \times 55 \times 3}$, and translation parameters $\gamma \in \mathbb{R}^{T \times 3}$. The optimization uses a differentiable surface vertex mapping function $\mathcal{S}(\beta, \theta, \gamma)$ and vertex normal function $\mathcal{N}(\beta, \theta_0, \gamma_0)$. For each frame, the latent marker $\Tilde{\mathbf{m}} \in \mathbb{R}^{T \times K \times 3}$ is calculated as
\begin{equation}
    \Tilde{\mathbf{m}}_i \equiv \mathcal{S}_{\mathcal{H}}(\beta, \theta_i, \gamma_i) + \mathbf{d} \mathcal{N}_{\mathcal{H}}(\beta, \theta_i, \gamma_i).
\end{equation}
We first select 12 frames to optimize and fix $\mathbf{d}$ and $\beta$, then optimize $\theta_i$, $\gamma_i$ for $i \in (1:T)$ by minimizing the loss terms based on $\| \Tilde{\mathbf{m}}_i - \mathbf{m}_i \|^2$, including Data Term, Surface Distance Energy, Marker Initialization Regularization, Pose and Shape Priors, Velocity Constancy, and Soft-Tissue Term (see supplementary materials). The overall objective function is the weighted sum of these terms to balance accuracy and plausibility.

\subsection{Body Parameters Refinement}
MoSh++ produces unnatural head shapes due to the fact that head markers were worn on a helmet and it sometimes produces unnatural finger poses.
Consequently, we refine body shape and pose parameters with three simple yet effective hard-coded physical rules. \textbf{i)} The neck and head length should approximate $1/7$ of the total length of the body. \textbf{ii)} Fingers, except for the thumb, should not bend backward. \textbf{iii)} We employ the Kolmogorov-Smirnov test, which reveals the data is similar to a normal distribution. Subsequently, we apply a Gaussian truncation approach, where all data points falling outside the $3\sigma$ range are adjusted to align with the $3\sigma$ threshold and blended with the adjacent 10 frames. We compare our refined body parameters (denoted as MoSh Refined) with Original BEAT, Retargeted SMPL-X, and MoSh SMPL-X in Figure \ref{fig:fig2}.

\begin{figure*}[]
\centerline{\includegraphics[width=2.1\columnwidth]{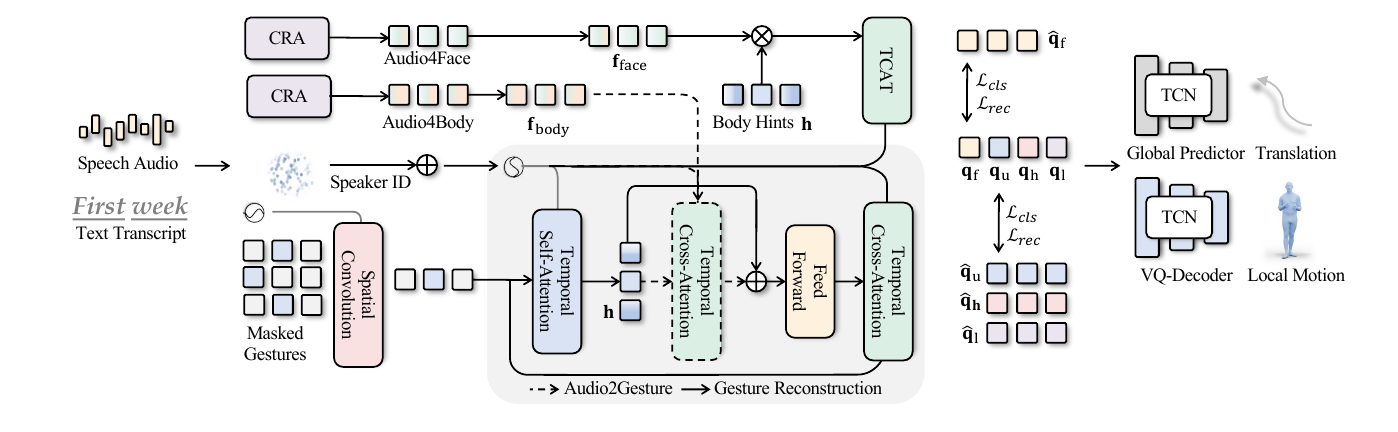}}
\vspace{-0.17in}
\caption{
\textbf{EMAGE} leverages two training paths: Masked Gesture Reconstruction (MG2G) and Audio-Conditioned Gesture Generation (A2G). The MG2G path focuses on encoding robust body hints through a spatial-temporal transformer gesture encoder and cross-attention gesture decoder. In contrast, the A2G path utilizes these body hints, and separated audio encoders to decode pretrained face and body latent features. A key component in this process is a switchable cross-attention layer, crucial for merging body hints and audio features. This fusion allows the features to be effectively disentangled and utilized for gesture decoding. Once the gesture latent features are reconstructed, EMAGE employs a pretrained VQ-Decoder to decode face and local body motions. Additionally, a pretrained Global Motion Predictor is used to estimate global body translations, further enhancing the model's capability to generate realistic and coherent gestures.         
}
\label{fig:fig3}
\end{figure*}


\begin{figure}
\centerline{
\includegraphics[width=1.0\columnwidth]{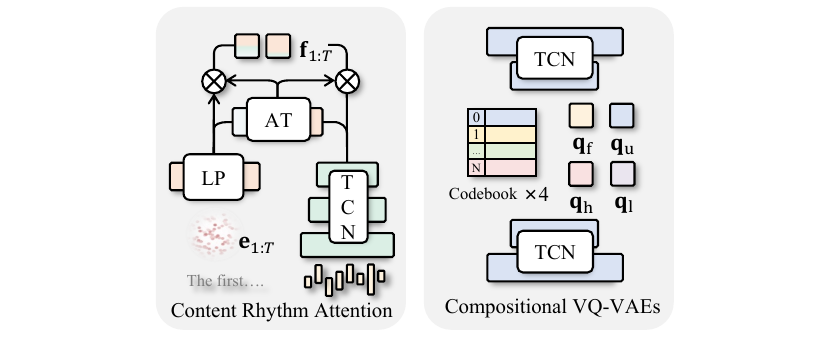}
}
\vspace{-0.07in}
\caption{
\textbf{Details of CRA and Pretrained VQ-VAEs.} {\textit{Left:} Content Rhythm Attention fuses speech rhythm (onset and amplitude) with content (pretrained word embeddings from text scripts) adaptively. This results in a preference for either content or rhythm in specific frames, which encourages the generation of semantical-aware gestures. \textit{Right:} We pretrain four compositional VQ-VAEs by reconstructing face, upper body, hands and lower body separately to disentangle audio-agnostic gestures explicitly.}}
\label{fig:fig4}
\end{figure}

\subsection{Blendshape Weights to FLAME Parameters}
Given the ARKit blendshape weights $\mathbf{b}_{\text{ARKit}} \in \mathbb{R}^{T \times 51}$, we aim to derive a transformation matrix $\mathbf{W} \in \mathbb{R}^{51 \times 103}$ that maps these into FLAME parameters $\mathbf{b}_{\text{FLAME}} \in \mathbb{R}^{T \times (100 + 3)}$, where 100 represents the number of dimensions for expression parameters and 3 for jaw movement. Due to discrepancies between the iPhone's and FLAME's template mesh topology, simply optimizing FLAME parameters by minimizing the geometric error between the ARKit template vertices and the wrapped FLAME vertices does not yield satisfactory results. Instead, we release a set of handcrafted blendshape templates $\mathbf{v}_{\text{template}} \in \mathbb{R}^{52}$ on FLAME following ARKit's FACS configuration. This approach allows direct driving of the FLAME topology vertices $\mathbf{v} \in \mathbb{R}^{T \times 5023 \times 3}$ from the given blendshape weights,  
\begin{equation}
\mathbf{v} = \mathbf{v}_{\text{template}}^0 + \sum_{j=1}^{51} \mathbf{b}_{\text{ARKit}, j} \cdot \mathbf{v}_{\text{template}}^j
\end{equation}
where $\mathbf{b}_{\text{ARKit}, j}$ is the weight of the $j$-th ARKit blendshape and $\mathbf{v}_{\text{template}}^j$ is the vertex position of the FLAME model template corresponding to the $j$-th blendshape. The term $\mathbf{v}_{\text{template}}^0$ represents the original template vertex position of the FLAME model. We optimize $\mathbf{W}$ by minimizing $\| \Tilde{\mathbf{v}}_j - \mathbf{v}_j \|_2$, where $\Tilde{\mathbf{v}}$ is obtained from FLAME's LBS $\mathcal{V}(\mathbf{b}_{\text{FLAME}})$. The comparisons between ARKit data, wrap-based, and our approach are shown in Figure \ref{fig:fig2}.

\section{EMAGE}
\label{sec:method}
We introduce the details of EMAGE, \textbf{E}xpressive \textbf{M}asked \textbf{A}udio-Conditioned \textbf{GE}sture Modeling (see Figure \ref{fig:fig3}). Given gestures $\mathbf{g} \in \mathbb{R}^{T \times (55 \times 6 +100+4+3)}$, representing 55 joints in Rot6D, $\mathbb{R}^{100}$ FLAME parameters, $\mathbb{R}^{4}$ foot contact labels, $\mathbb{R}^{3}$ global translations, and speech audio $\mathbf{a} \in \mathbb{R}^{T \times sk}$, where $sk = sr_{\text{audio}}/fps_{\text{gestures}}$, EMAGE jointly optimizes masked gesture reconstruction and audio-conditioned gesture generation. This optimization enhances performance during inference and enables the use of partially masked gestures to complete holistic gestures. 
To achieve this, we first model the quantized latent space in Sec.~\ref{sec4.1}, following \cite{talkshow:yi2022generating, xing2023codetalker}. We then design a separated speech audio encoder using Content Rhythm Self-Attention (CRA) as detailed in Sec.~\ref{sec4.2}. Subsequently, we learn the body hints from masked gestures via the Masked Audio Gesture Transformer, explained in Sec.~\ref{sec4.3}. Finally, the decoding strategy for different body segments is discussed in Sec.~\ref{sec4.4}.  

\subsection{Compositional Discrete Face and Body Prior}
\label{sec4.1}
We model full-body gestures in the separated quantized latent space (Figure \ref{fig:fig4}) for several reasons. Similar to \cite{talkshow:yi2022generating}, the body and hands improves the diversity of the results. However, we additionally separate the face and lower body due to their differing correlations with audio, \textit{i.e.,} using a single VQ-VAE to encode both the upper and lower body may lead the model to overlook gestures that occur less frequently - regardless of their relation to the audio. Specifically, a single model might focus only on recovering lower body motion, neglecting elbow movements in a speaker who is constantly walking during the conversation.

The separated quantized latent space $Q =  \{\mathbf{q}_\text{f}, \mathbf{q}_u, \mathbf{q}_\text{h}, \mathbf{q}_\text{l}\}$ for the Face $\mathbf{g}_\text{f} \in \mathbb{R}^{T \times 106}$, Upper body $\mathbf{g}_\text{u} \in \mathbb{R}^{T \times78}$, Hands $\mathbf{g}_\text{h} \in \mathbb{R}^{T \times 180}$, and Lower body $\mathbf{g}_\text{l} \in \mathbb{R}^{T \times (54+4)}$ are from four VQ-VAEs. 
The VQ-VAE for each is optimized by jointly optimizing the following loss terms,
\begin{equation}
q_i= \argmin_{q_i \in \mathbf{q}} \|z_j - q_i\|^2
\end{equation}
\begin{align}
\mathcal{L}_\text{VQ-VAE} =\ &\mathcal{L}_{rec}(\mathbf{g}, \hat{\mathbf{g}}) + \mathcal{L}_{\text{vel}}(\mathbf{g}', \hat{\mathbf{g}}') + \mathcal{L}_{\text{acc}}(\mathbf{g}'', \hat{\mathbf{g}}'') \nonumber \\
&+ \|\text{sg}[\mathbf{z}] - \mathbf{q}\|^2 + \|\mathbf{z} - \text{sg}[\mathbf{q}]\|^2,
\end{align}
where $\mathbf{z}$ is the encoded $\mathbf{g}$ with a temporal window size $w = 1$. $\mathcal{L}_{\text{rec}}$, $\mathcal{L}_{\text{vel}}$, and $\mathcal{L}_{\text{acc}}$ are Geodesic \cite{tykkala2011direct} and L1 losses, $\text{sg}$ is a stop gradient operation, we set the weight of the commitment loss \cite{van2017neural} (the last term) to 1 in this paper.

\subsection{Content and Rhythm Self-Attention}
\label{sec4.2}
Given the speech audio, $\mathbf{s}$, inspired by \cite{Ao2023GestureDiffuCLIP}, we employ onset $\mathbf{o}$ and amplitude $\mathbf{a}$ as explicit audio rhythm, alongside the pretrained embeddings \cite{bojanowski2017enriching} $\mathbf{e}$ from transcripts as content. Different from previous approaches, which typically add the rhythm and content features, we leverage self-attention to merge these features adaptively. This approach is driven by the observation that for specific frames, the gestures are more related to content (semantic-aware) or rhythm (beat-aware). The rhythm and content features are first encoded into time-aligned features $\mathbf{r}_{1:T}$ and $\mathbf{c}_{1:T}$, using a Temporal Convolutional Network (TCN) and linear mapping, respectively. For each time step $t \in \{1, \dots, T\}$, we merge the rhythm and content features by:
\begin{equation}
\begin{aligned}
\mathbf{f}_{1:T} &= \alpha \times \mathbf{r}_{1:T} + (1 - \alpha) \times \mathbf{c}_{1:T}, \\
\alpha &= \text{Softmax}(\mathcal{AT}(\mathbf{r}_{1:T}, \mathbf{c}_{1:T})),
\end{aligned}
\end{equation}
where $\mathcal{AT}$ is a 2-layer MLP. We apply two separate CRA encoders for the face and body.

\begin{figure}
\centerline{
\includegraphics[width=1.0\columnwidth]{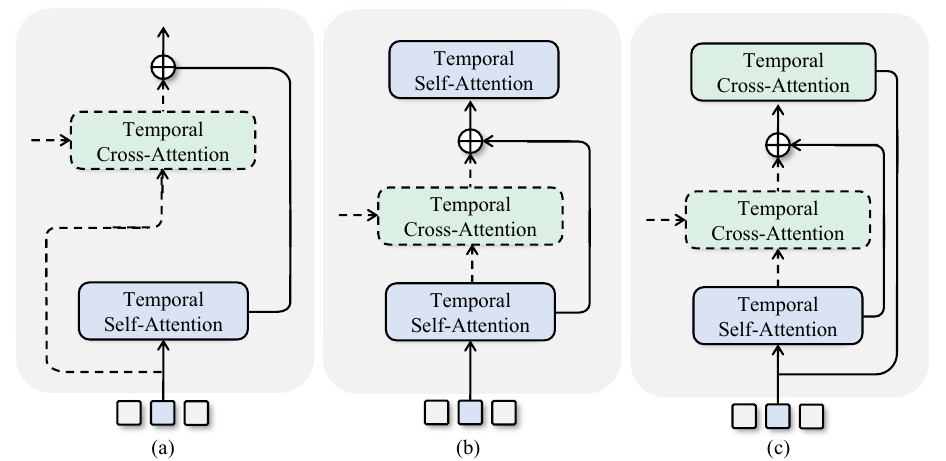}
}
\vspace{-0.08in}
\caption{
\textbf{Comparison of Forward Path Designs.} Straightforward fusion module (a) merges audio features without refined body features and recombines audio features based only on position embedding. The Self-Attention decoder module (b), adopted in previous MLM models \cite{devlin2018bert, lan2019albert}, is limited for tasks requiring auto-regressive inference. Our design (c) considers effective audio feature fusion and auto-regressive decoding.}
\label{fig:fig42}
\end{figure}

\subsection{Masked Audio-Conditioned Gesture Modeling}
\label{sec4.3}
We propose a Masked Audio Gesture Transformer to leverage different training paths (see Fig.~\ref{fig:fig42} for the motivation behind the architecture designs). Given temporal and spatially masked gestures $\overline{\mathbf{g}} \in \mathbb{R}^{T \times 337}$, we first replace the masked tokens with learned masked embeddings $e_{\text{mask}} \in \mathbb{R}^{256}$, as the value zero still represents specific motion content, e.g., a T-pose. We linearly increase the ratio of masked joints and frames from 0 to 95\% according to the training epochs.

A spatial convolutional $\mathcal{SC}$ encoder initially summarizes the spatial information, and compress the spatial feature into $\mathbb{R}^{T \times 256}$. We then employ a temporal self-attention (without feed-forward) $\mathcal{TSA}$ to refine the summarized spatial features 
\begin{equation}
\mathbf{h} = \mathcal{TSA}(\mathcal{SC}(\overline{\mathbf{g}}) + \mathbf{p}_t),
\end{equation}
where $\mathbf{h} \in \mathbb{R}^{T \times 512}$ represents encoded body hints, and $\mathbf{p}_t$ is the sum of a learned speaker embedding and PPE \cite{faceformer2022}. Then, a straightforward temporal cross-attention Transformer decoder $\mathcal{TCAT}$ is adopted for the reconstructed latent $\hat{\mathbf{q}}_{\text{mg2g}}$,
\begin{equation}
\hat{\mathbf{q}}_{\text{mg2g}} = \mathcal{TCAT}(\mathbf{h} + \mathbf{p}_t).
\end{equation}
We minimize the L1 distance in the latent space,
\begin{equation}
    \mathcal{L}_{\text{mg2g}} = \|\hat{\mathbf{q}}_{\text{mg2g}} - \mathbf{q}\|.
\end{equation}

Masked gesture reconstruction encodes effective body hints, with the key being to leverage these body hints for gesture generation. We employ a selective fusion for audio and body hints via a temporal cross-attention $\mathcal{TCA}$. Subsequently, we use the merged audio-gesture features for audio-conditioned gesture latent reconstruction,
\begin{equation}
\hat{\mathbf{q}}_{\text{a2g}} = \mathcal{TCAT}(\mathcal{TCA}(\mathbf{h} + \mathbf{p}_t, \mathbf{f}_{\text{body}}), \overline{\mathbf{g}} + \mathbf{p}_t).
\end{equation}
We optimize both the latent code class classification cross-entropy loss $\mathcal{L}_{\text{a2g-rec}}$ and the latent reconstruction loss $\mathcal{L}_{\text{a2g-cls}}$ to encourage diverse results.

\subsection{Face and Translation Decoding}
\label{sec4.4}
Considering that the face is weakly related to body motion, applying the same operation – recombining the audio features based on body hints – is not reasonable. Therefore, we directly concatenate the masked body hints with the audio features for the final decoding of facial latent,
\begin{equation}
\hat{\mathbf{q}}_{\text{f}} = \mathcal{TCAT}(\mathbf{f}_{\text{face}} \oplus \mathbf{h}, \mathbf{p}_t). 
\end{equation}
Once we have obtained the local lower body motion $\tilde{\mathbf{g}}_l \in \mathbb{R}^{T \times (54+4)}$ from the VQ-Decoder, we estimate the global translations $\tilde{\mathbf{t}} \in \mathbb{R}^{T \times 3}$ with a pretrained Global Motion Predictor $\tilde{\mathbf{t}} = \mathcal{G}(\tilde{\mathbf{g}}_l)$, which significantly reduces foot sliding.

\begin{table}
\centering
\resizebox{0.90\linewidth}{!}{
\begin{tabular}{lcccc}
         & Top.-B & Top.-F & Body        & Face         \\ 
\hline
VOCA\cite{VOCA2019}     & -      & FLAME  & -           & \textbf{38.3~± 5.63}  \\
AMASS\cite{AMASS:ICCV:2019}    & SMPL-X  & FLAME  & 42.0~± 3.60 & -            \\
TalkSHOW\cite{talkshow:yi2022generating} & SMPL-X  & SMPL-X  & 14.4~± 2.19 & 26.1~± 6.42  \\
BEAT2 (Ours)    & SMPL-X  & FLAME  & \textbf{43.6~± 3.38} & 35.7 ± 5.91 
\end{tabular}}
\vspace{-0.08in}
\caption{\textbf{User Preference Win Rate between Existing Datasets.} `Top.' denotes the topology of the mesh. The results show our BEAT2 dataset outperforms the existing PGT dataset \cite{talkshow:yi2022generating} in both body and face aspects. It also performs slightly better on body and lower on face when compared with previous mocap \cite{AMASS:ICCV:2019} and 3D scan-based datasets \cite{VOCA2019}, respectively.}
\label{tab:tab3}
\vspace{-0.2cm}
\end{table}

\begin{figure*}[]
\begin{center}
\includegraphics[width=2\columnwidth]{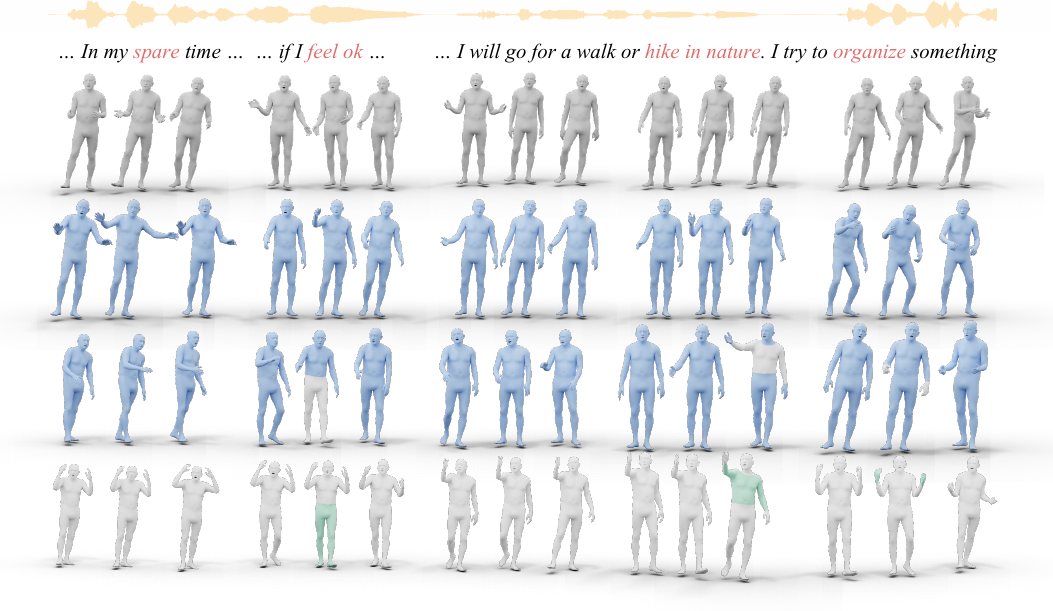}
\vspace{-0.6cm}
\caption{
\textbf{Subjective Examples.} \textit{Top to down}: GroundTruth, \textcolor{Teal}{Generated W/O Body Hints}, \textcolor{Teal}{Generated With} \textcolor{gray}{Body Hints}, \textcolor{MyGr}{Visible} \textcolor{gray}{Body Hints}, EMAGE generates diverse, semantic-aware and audio synchronized body gestures, \textit{e.g.} raise the both hands for ``spare time'', relaxing gestures for ``hike in nature''. Besides, as shown in the third and bottom rows, EMAGE is flexible to accept non-audio-synchronized body hints on any frames and joints, to guide or customize the generated gestures explicitly, \textit{ e.g.}, repeat a similar motion for raising hands, change the walking orientations, \textit{etc.} Note that in this figure, The gray joints of generated results are not copies of visible hints. 
}
\vspace{-0.4cm}
\label{fig:fig5}
\end{center}
\end{figure*}
\section{Experiments}
We separate the evaluation into two categories: dataset quality and model ability. After removing data with low finger quality from BEAT, BEAT2 is reduced to 60 hours. We further split it into BEAT2-Standard (27 hours) and BEAT2-Additional (33 hours), based on the type of speech and conversation sections \cite{liu2022beat}. While acted gestures in speech sections are more diverse and expressive, spontaneous gestures in conversational sections tend to be more natural yet less varied. We report the results on BEAT2-Standard Speaker-2 with an 85\%/7.5\%/7.5\% train/val/test split.

\subsection{Dataset Quality Evaluation}
We compare our dataset with the state-of-the-art pseudo ground truth (PGT) dataset, TalkSHOW \cite{talkshow:yi2022generating}, for both face and body. Additionally, we compare it with the AMASS \cite{AMASS:ICCV:2019} dataset for the body and VOCA \cite{VOCA2019} for the face as references. Due to the varying lengths of sequences in each dataset, we sample 100 comparison pairs, each with an equal duration ranging from 2 to 4 seconds. 
In a perceptual study, each participant evaluates a random set of 40 pairs during a 10-minute session by selecting the sequence they consider to have the best captured quality. In total, 60 participants were invited. It is important to note that participants are instructed to compare only the upper body results, as TalkSHOW contains only the upper body. The results are shown in Table \ref{tab:tab3}.

\subsection{Model Ability Evaluation}
We report results using BEATv1.3 and BEATv2. The latter  has facial data refined by animators and hand data filtered by annotators. 

\begin{table}[t]
\centering
\resizebox{0.99\linewidth}{!}{
\begin{tabular}{lccccc}
                 & FGD $\downarrow$ & BC $\uparrow$ & Diversity~$\uparrow$ & MSE ~$\downarrow$ & LVD ~$\downarrow$  \\ 
\midrule
FaceFormer\cite{faceformer2022}       & -                      & -                    & -                    &  7.787                       & 7.593                         \\
CodeTalker\cite{xing2023codetalker}       & -                      & -                    & -                    & 8.026                  & 7.766                         \\
S2G\cite{ginosar2019learning}              & 28.15                 & 4.683      & 5.971               & -                       & -                        \\
Trimodal\cite{yoon2020speech}         & 12.41                 & 5.933                & 7.724                & -                       & -                        \\
HA2G\cite{ha2g:liu2022learning}             & 12.32                 & 6.779                & 8.626                & -                       & -                        \\
DisCo\cite{liu2022disco}            & 9.417                 & 6.439                & 9.912               & -                       & -                        \\
CaMN\cite{liu2022beat}             & 6.644                  & 6.769               & 10.86                & -                       & -                        \\
DiffStyleGesture\cite{yang2023diffusestylegesture} & 8.811                  & 7.241                & 11.49                & -                       & -                      \\
Habibie \textit{et al}.\cite{habibie2021learning}   & 9.040                & 7.716                & 8.213                & 8.614                   & 8.043                    \\
TalkSHOW\cite{talkshow:yi2022generating}         & 6.209                  & 6.947                & \textbf{13.47}                & 7.791                  & 7.771                  \\
EMAGE (Ours)      & \textbf{5.512}                  & \textbf{7.724}                & 13.06              & \textbf{7.680}                   & \textbf{7.556}                   
\end{tabular}}
\vspace{-0.08in}
\caption{\textbf{Quantitative evaluation on BEATv2.} We report FGD $\times 10^{-1}$, BC $\times 10^{-1}$, Diversity, MSE $\times 10^{-8}$, and LVD $\times 10^{-5}$. For body gestures, EMAGE significantly improves the FGD, indicating that the generated results are closer to the GT. This shows the effectiveness of body hints from masked gesture modeling.}
\label{tab:tab4}
\vspace{-0.2cm}
\end{table}

\begin{table}[tbh]
\centering
\resizebox{0.85\linewidth}{!}{
\begin{tabular}{lccc}
               & Holistic             & Body                 & Face                  \\ 
\hline
Habibie \textit{et al}.\cite{habibie2021learning} & 12.4~± 3.70          & 15.9 ± 6.49          & 10.8~± 3.19            \\
TalkSHOW\cite{talkshow:yi2022generating}       & 34.9~± 5.79          & 40.4 ± 8.22          &  33.2 ± 6.03         \\
EMAGE (Ours)    & \textbf{52.7~± 7.91} & \textbf{44.7 ± 8.68} & \textbf{56.0 ± 7.80} 
\end{tabular}}
\vspace{-0.08in}
\caption{\textbf{User Preference Win Rate On Generated Results.} The results indicate that our generated outcomes are perceived as more realistic and believable, with a 14\% and 23\% higher user preference for body and face gestures, respectively.}
\label{tab:tab5}
\vspace{-0.2cm}
\end{table}

\smallskip
\noindent\textbf{Metrics.} We adopt \textbf{FGD}\cite{yoon2020speech} to evaluate the realism of the body gestures. Then, we measure \textbf{Diversity}\cite{li2021audio2gestures} by calculating the average L1 distance between multiple body gesture clips, and use \textbf{BC} \cite{li2021ai} to assess speech-motion synchrony. For the face, we calculate the vertex \textbf{MSE}\cite{xing2023codetalker} to measure the positional distance and the vertex L1 difference \textbf{LVD} \cite{talkshow:yi2022generating} between the GT and generated facial vertices.

\smallskip
\noindent\textbf{Compared Methods.} We first compare our method with representative state-of-the-art methods in body gesture generation \cite{ginosar2019learning, yoon2020speech, ha2g:liu2022learning, liu2022disco, liu2022beat, yang2023diffusestylegesture} and talking head generation \cite{faceformer2022, xing2023codetalker} by reproducing their methods for body and face, respectively. In addition, we reproduce two previous state-of-the-art holistic pipelines, Habibie \textit{et al.}~\cite{habibie2021learning} and TalkSHOW \cite{talkshow:yi2022generating}, whose original implementations are limited to the upper body. We add a lower-body decoder to Habibie {\em et al.}~and a lower-body VQ-VAE to TalkSHOW.

\vspace{-0.3cm}
\subsubsection{Quantitative and Qualitative Analysis}
\begin{figure}
\centerline{
\includegraphics[width=\columnwidth]{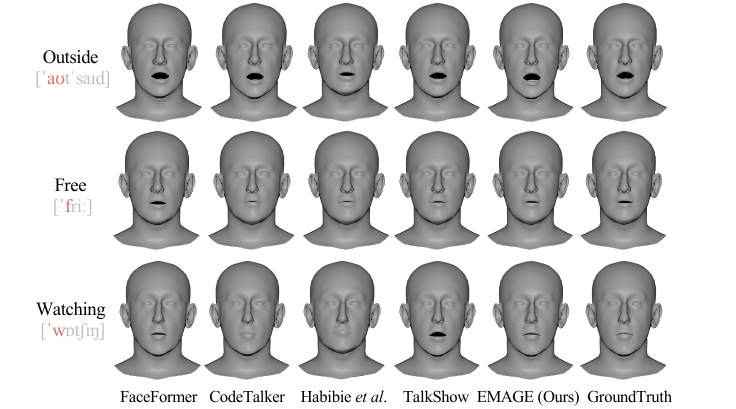}
}
\vspace{-0.2cm}
\caption{
\textbf{Comparison of Generated Facial Movements.} 
Compared with previous state-of-the-art talking face generation methods FaceFormer \cite{faceformer2022} and CodeTalker \cite{xing2023codetalker} as well as holistic gesture generation methods Habibie \textit{et al.}~\cite{habibie2021learning} and TalkSHOW  \cite{talkshow:yi2022generating}.
Note that CodeTalker has higher MSE than EMAGE on BEATv2 (Table \ref{tab:tab4}, lower is better) but is subjectively realistic. 
EMAGE gets good lip motions by leveraging both the face model and masked body hints.
}
\label{fig:fig6}
\vspace{-0.4cm}
\end{figure}

\vspace{-0.2cm}
As shown in Table \ref{tab:tab4}, with a four-frame seed pose, our method outperforms previous SOTA algorithms. For qualitative results, see Figures \ref{fig:fig5} and \ref{fig:fig6}. Furthermore, we conduct a perceptual study. Maintaining the same setup of 60 participants, each participant evaluates 40 pairs of 10-second results do decide which is most believable; this gives the win rate shown in Table \ref{tab:tab5}.
%
%


\subsubsection{Ablation Analysis}
\noindent\textbf{Performance of Baseline.}
As shown in Table \ref{tab:tab6}, we start with a teacher-force transformer-based baseline. This baseline, inspired by FaceFormer \cite{faceformer2022}, adopts a 1-layer cross-attention transformer decoder and replaces the audio features from Wav2Vec2 \cite{baevski2020wav2vec} with our custom TCN \cite{bai1803empirical} and trainable word embeddings \cite{bojanowski2017enriching}.

\noindent\textbf{Effect of Multiple VQ-VAEs.}
Simply applying one VQ-VAE \cite{van2017neural, guo2022tm2t} for full-body movements, including the face, decreases performance in facial movements. This is because the VQ-VAE is trained to minimize the average loss for the full body, and some speakers' most frequent movements are not related to the audio. Implementing separated VQ-VAEs allows the model to better leverage the advantages of discrete priors.

\noindent\textbf{Effect of Content Rhythm Self-Attention.}
Adaptive fusion of rhythm and content features shows improvement in both FGD and Alignment. It selectively focuses the current motion more on rhythm or content features based on the training data's distribution. Moreover, we observed more semantic-aware results when applying CRA.

\noindent\textbf{Effect of Masked Body Gesture Hints.}
The improvement across all objective metrics demonstrates that our model effectively leverages spatial-temporal gesture priors, reducing the likelihood of incorrect gesture sampling. Furthermore, masked gesture modeling is key to enabling the network to accept predefined gestures in specific frames.

\begin{table}
\centering
\resizebox{0.99\linewidth}{!}{
\begin{tabular}{lccccc}
               & FGD $\downarrow$ & BC~$\uparrow$ & Diversity $\uparrow$ & MSE~$\downarrow$ & LVD~$\downarrow$  \\ 
\midrule
Ground Truth   & 0                & 6.896             & 13.074               & 0                & 0                 \\
Reconstruction & 3.913            & 6.758             & 13.145               & 0.841            & 6.389             \\
Baseline       & 13.080           & \textbf{6.941}    & 8.3145               & 1.442            & 9.317             \\
+ VQVAE        & 9.787            & 6.673             & 10.624               & 1.619            & 9.473             \\
+ 4 VQVAE      & 7.397            & 6.698             & 12.544               & 1.243            & 8.938             \\
+ CRA          & 6.833            & 6.783             & 12.676               & 1.186            & 8.715             \\
+ Masked Hints & \textbf{5.423}   & 6.794             & \textbf{13.057}      & \textbf{1.180}   & \textbf{9.015}   
\end{tabular}}
\vspace{-0.08in}
\caption{\textbf{Abliation Analysis on BEATv1.3}}
\label{tab:tab6}
\end{table}

\begin{table}
\centering
\resizebox{0.90\linewidth}{!}{
\begin{tabular}{lcccccc}
          & body & hands & face & FGD $\downarrow$ & BC~$\uparrow$ & Diversity $\uparrow$  \\ 
\midrule
BESTX     &  \checkmark    &  \checkmark     &  \checkmark    & 5.423            & 6.794            & 13.057                 \\
+ Trinity\cite{ferstl2018investigating} &  \checkmark    &       &      & 5.319            & 6.843            & 13.346                 \\
+ AMASS\cite{AMASS:ICCV:2019}   & \checkmark     & \checkmark      &      & 5.174            & 6.769            & 14.318                     
\end{tabular}}
\vspace{-0.08in}
\caption{\textbf{Training EMAGE on Multiple Datasets.} EMAGE demonstrates flexibility by training on multiple datasets, even when only a subset of holistic gestures is available. This approach further improves the objective metrics on the BEATv1.3 test set.}
\vspace{-0.3cm}
\label{tab:tab7}
\end{table}

\vspace{-0.3cm}
\subsubsection{Ability of Multi-Dataset Training.}
We demonstrate that EMAGE can effectively combine multiple non-holistic datasets for training by jointly training with the Trinity \cite{ferstl2018investigating} and AMASS \cite{AMASS:ICCV:2019} datasets, by using only the upper body and audio pairs from Trinity and only body and hands from AMASS. Separate VQ-VAEs are trained for BEAT2, Trinity, and AMASS, with separate MLP heads implemented for codebook classification. The results in Table \ref{tab:tab7} show that incorporating data improves performance on the BEATv1.3 test set.

\vspace{-0.2cm}
\section{Conclusion}
In this work, we present EMAGE, a framework designed to accept partial gestures as input for completing audio-synchronized holistic gestures. It demonstrates that leveraging masked gesture reconstruction can significantly enhance the performance of audio-conditioned gesture generation. Furthermore, the design of EMAGE enables the training on multiple datasets, further improving performance. Alongside EMAGE, we release BEAT2, the largest multi-modal gesture dataset consistent with SMPL-X and FLAME. We hope that BEAT2 will contribute to knowledge and model sharing across various subareas.

\noindent\textbf{Disclosures:}
While You Zhou, Xuefei Zhe, Naoya Iwamoto, and Bo Zheng are employees of Huawei Tokyo Research Center, this work was done on their own time with the approval of their employer.
MJB CoI: \url{https://files.is.tue.mpg.de/black/CoI_CVPR_2024.txt}

{
    \small
    \bibliographystyle{ieeenat_fullname}
    \bibliography{macros,main}
}

\appendix

\setcounter{page}{1}

\twocolumn[
\centering
\Large
\textbf{EMAGE: Towards Unified Holistic Co-Speech Gesture Generation via Expressive Masked Audio Gesture Modeling} \\
\vspace{0.5em}Supplementary Material \\
\vspace{1.0em}]

\appendix
\noindent This supplemental document contains seven sections: 
\begin{itemize}[leftmargin=*]
\setlength\itemsep{-.3em}

\item Evlauation Metrics (Section \ref{secsup1}).

\item BEAT2 Dataset Details (Section \ref{secsup2}).

\item Baselines Reproduction Details (Section \ref{secsup3}).

\item Settings of EMAGE (Section \ref{secsup4}).

\item Visualization Blender Addon (Section \ref{secsup5}).

\item Training time (Section \ref{secsup6}). 

\item Importance of lower body motion (Section \ref{secsup7}). 
\end{itemize}










\section{Evaluation Metrics}
\label{secsup1}
\paragraph{Fr\'{e}chet Gesture Distance} A lower FGD, as referenced by \cite{yoon2020speech}, indicates that the distribution between the ground truth and generated body gestures is closer. Similar to the perceptual loss used in image generation tasks, FGD is calculated based on latent features extracted by a pretrained network,
\begin{equation}
\label{eqfid}
\resizebox{.85\hsize}{!}{$
\operatorname{FGD}(\mathbf{g}, \hat{\mathbf{g}})=\left\|\mu_{r}-\mu_{g}\right\|^{2}+\operatorname{Tr}\left(\Sigma_{r}+\Sigma_{g}-2\left(\Sigma_{r} \Sigma_{g}\right)^{1 / 2}\right),$}
\end{equation}
where $\mu_{r}$ and $\Sigma_{r}$ represent the first and second moments of the latent features distribution $z_{r}$ of real human gestures $\mathbf{g}$, and $\mu_{g}$ and $\Sigma_{g}$ represent the first and second moment of the latent features distribution $z_{g}$ of generated gestures $\hat{\mathbf{g}}$. We use a Skeleton CNN (SKCNN) based encoder \cite{aberman2020skeleton}  and a Full CNN-based decoder as the autoencoder pretrained network. The network is pretrained on both BEAT2-Standard and BEAT2-Additional Data.  The choice of SKCNN over a Full CNN encoder is due to its enhanced capability in capturing gesture features, as indicated by a lower reconstruction MSE loss (0.095 compared to 0.103).

\paragraph{L1 Diversity} A higher Diversity \cite{li2021audio2gestures} indicates a larger variance in the given gesture clips. We calculate the average L1 distance from different $N$ motion clips as follows:
\begin{equation}
\resizebox{.75\hsize}{!}{$
    \text{L1 div.} =  \frac{1}{2 N (N-1)} \sum_{t=1}^{N} \sum_{j=1}^{N} \left\|p_{t}^{i}-\hat{p}_{t}^{j}\right\|_{1},$}
\end{equation} 
where $p_{t}$ represents the position of joints in frame $t$. We calculate diversity on the entire test dataset. Additionally, to compute joint positions, the translation is set to zero, implying that L1 Diversity is focused solely on local motion. 

\paragraph{Beat Constancy (BC)} A higher BC, as defined by \cite{li2021ai}, suggests a closer alignment between the gesture's rhythm and the beat of the audio. We identify the beginning of speech as the audio beat and consider the local minima of the velocity of the upper body joints (excluding fingers) as the motion beat. The synchronization between audio and gesture is computed in the following manner:
\begin{equation}
\label{align}
\resizebox{.80\hsize}{!}{$
\text{BC}= \frac{1}{g} \sum_{b_{g}\in g} \exp \left(-\frac{\min _{b_{a}\in a}\left\|b_{g}-b_{a}\right\|^{2}}{2 \sigma^{2}}\right),$}
\end{equation}
where $g$ and $a$ represent the sets of gesture beat and audio beat, respectively.

\section{BEAT2 Dataset Details}
\label{secsup2}
\paragraph{Statistics} The original BEAT dataset, as described by \cite{liu2022beat}, contains 76 hours of data for 30 speakers. We exclude speakers 8, 14, 19, 23, and 29, which account for 16 hours of data, due to noise in the finger data, leaving 60 hours of data for 25 speakers (12 female and 13 male). The speech and conversation portions are categorized into BEAT2-standard and BEAT2-additional, containing 27 and 33 hours respectively.  We adopt an 85\%, 7.5\%, and 7.5\% split for BEAT2-standard, maintaining the same ratio for each speaker. BEAT2-additional is utilized for further improving the network's robustness. The results presented in this paper are based on training with BEAT2-standard speaker-2 only. The dataset includes 1762 sequences with an average length of 65.66 seconds per sequence. Each recording in a sequence is a continuous answer to a daily question. Additionally, we report a comparison between TalkShow \cite{talkshow:yi2022generating} and our dataset in terms of Diversity and Beat Constancy (BC), as shown in Table \ref{tab:tab8}.

\begin{table}[h]
\centering
\caption{\textbf{Diversity and BC Comparisons.} The local and global diversity refers to the variance in joint positions with and without global translations, respectively.}
\label{tab:tab8}
\resizebox{0.80\linewidth}{!}{
\begin{tabular}{lccc}
         & BC~$\uparrow$ & Diversity-L~$\uparrow$ & Diversity-G~$\uparrow$  \\ 
\midrule
TalkShow \cite{talkshow:yi2022generating} & 6.104         & 5.273                  & 5.273                     \\
BEAT2 (Ours)    & 6.896         & 13.074                 & 27.541                   
\end{tabular}}
\end{table}

\paragraph{Loss Terms of MoSh++} MoSh's optimization involves loss functions including a Data Term, Surface Distance Energy, Marker Initialization Regularization, Pose and Shape Priors, and a Velocity Constancy Term,  which are described as follows:
\begin{itemize}[leftmargin=*]
\setlength\itemsep{-.3em}

\item Data Term ($E_D$): Minimizing the squared distance between simulated and observed markers. In the given context, the $\tilde{M}, \beta, \Theta$, and $\Gamma$ represent the latent markers, body shape, poses, and body location respectively:
\begin{equation}
E_D(\tilde{M}, \beta, \Theta, \Gamma) = \sum_{i,t} ||\hat{m}(\tilde{m}_i, \beta, \theta_t, \gamma_t) - m_{i,t}||^2.
\end{equation}

\item Surface Distance Energy ($E_S$): Ensuring markers maintain prescribed distances from the body surface:
\begin{equation}
E_S(\beta, \tilde{M}) = \sum_{i} ||r(\tilde{m}_i, S(\beta, \theta_0, \gamma_0)) - d_i||^2.
\end{equation}

\item Marker Initialization Regularization ($E_I$): Penalizing deviations of estimated markers from initial positions:
\begin{equation}
E_I(\beta, \tilde{M}) = \sum_{i} ||\tilde{m}_i - v_i(\beta)||^2.
\end{equation}

\item Pose and Shape Priors: Penalizing deviations from mean shape and pose:
\begin{equation}
E_\beta(\beta) = (\beta - \mu_\beta)^T \Sigma^{-1}_\beta (\beta - \mu_\beta),
\end{equation}
\begin{equation}
E_\theta(\Theta) = \sum_{t} (\theta_t - \mu_\theta)^T \Sigma^{-1}_\theta (\theta_t - \mu_\theta).
\end{equation}

\item Velocity Constancy Term ($E_u$): Reducing marker noise and ensuring movement consistency:
\begin{equation}
E_u(\Theta) = \sum_{t=2}^{n} ||\theta_t - 2\theta_{t-1} + \theta_{t-2}||^2.
\end{equation}

\end{itemize}
The overall objective function is the weighted sum of these terms, balancing accuracy and plausibility:
\begin{equation}
E(\tilde{M}, \beta, \Theta, \Gamma) = \sum_{\omega \in \{D, S, \theta, \beta, I, u\}} \lambda_{\omega} E_{\omega}(\cdot).
\end{equation}

More details and pseudo code of the head and neck shape optimization are available in the code release.

\paragraph{Details of FLAME Parameter Optimization} To animate a face using the SMPL-X model with ARKit parameters from the BEAT dataset, we estimate FLAME expression parameters by minimizing the geometric error between an animated ARKit and FLAME model. Addressing the optimization challenges posed by differing mesh structures, we construct an ARKit-compatible FLAME model utilizing Faceit, a Blender add-on tailored for crafting ARKit blendshapes. By driving the ARKit-aligned FLAME model with each set of ARKit parameters from the BEAT dataset, we obtain original FLAME expression parameters by minimizing the L2 distance loss between equivalent vertices. Finally, the optimized FLAME expression parameters can be directly applied to SMPL-X. For facial identity parameters, we preserve the same identity parameters on SMPL-X after body fitting with MoSh++ \cite{AMASS:ICCV:2019}.

\section{Baselines Reproduction Details}
\label{secsup3}
\paragraph{Number of Joints} All baseline methods output full-body joint rotations represented by  $\mathbf{g} \in \mathbb{R}^{T \times (55 \times 6)}$ and, in addition to rotations, they decode global translations $\in \mathbb{R}^{T \times 3}$. To provide a thorough comparison, we present subjective results for both the upper body (excluding global motion) and the full body.

\paragraph{Autoregressive Training}
We observe that autoregressive training/inference-based models, such as FaceFormer and CodeTalker \cite{faceformer2022, xing2023codetalker}, perform worse than non-autoregressive methods. In non-autoregressive settings, only positional embedding is used as input for cross-attention to audio features, particularly when training with Rot6D and axis-angle representations. The network architecture of FaceFormer and CodeTalker is based on transformers and was initially proposed for training with the representation of vertex offsets. As shown in Table \ref{tab:tab9}, we find that non-autoregressive training improves performance with FLAME's parameters. The results in this paper are obtained using a non-autoregressive training approach. Non-autoregressive training techniques have also been employed in the training of EMAGE.

\begin{table}[h]
\centering
\caption{\textbf{Vertex Errors (MSE) with Different Training Methods.} `FF' and `CT' refer to FaceFormer \cite{faceformer2022} and CodeTalker \cite{xing2023codetalker}, respectively. `TF', `AR', and `NonAR' represent Teacher-Force, AutoRegressive, and Non-AutoRegressive training, respectively. We train on the VOCA dataset with a vertex loss, and BEAT2 with a FLAME parameter loss combined with a vertex loss. Results indicate that the same method performs differently when using the two representations; in BEAT2, non-autoregressive training demonstrates superior performance. The average MSE is calculated on 5023 and 10475 vertices for VOCA and BEAT2, respectively:
}
\label{tab:tab9}
\resizebox{0.95\linewidth}{!}{
\begin{tabular}{lcccccc}
              & FF-TF & FF-AR & FF-NonAR & CT-TF & CT-AR & CT-NonAR  \\ 
\hline
VOCA (x10-7)  & 6.636     & 6.023     & 6.138     &  7.914     & 7.637      &   7.541       \\
BEAT2 (x10-7) & 2.167 & 3.704 & 1.195    & 2.079 & 4.120 & 1.243    
\end{tabular}}
\end{table}

\paragraph{Adversarial Training} We omit the adversarial training in Speech2Gesture \cite{ginosar2019learning}, CaMN \cite{liu2022beat}, and Habibie \textit{et al} \cite{habibie2021learning}, because their outputs with adversarial training show noticeable jitter, even when we increase the weight for the velocity loss. Similar effects are also observed in training with 3D data for Speech2Gesture \cite{ginosar2019learning}, as reported in the study by \cite{li2021audio2gestures}.

\paragraph{Lower Body VQ-VAE for TalkShow} We introduce an additional VQ-VAE for TalkShow, utilizing their autoregressive (AR) model to jointly predict the class index of the upper body, hands, and lower body. The global translations are encoded in conjunction with lower body joints.

\section{Settings of EMAGE}
\label{secsup4}
\paragraph{Training} We train our method for 400 epochs, gradually increasing the ratio of masked joints from 0 to 95\% linearly according to the training epoch. This approach proves more effective than a fixed masked ratio, such as 25\%, based on our experiments. The learning rate is  2.5$e$-4, and we use the Adam optimizer with a gradient norm clipped at 0.99 to ensure stable training.

\paragraph{Structure of VQ-VAE} We employ the same CNN-based VQ-VAE \cite{guo2022tm2t} for all four body segments. The downsample rate is set to 1 to achieve the best reconstruction quality. We utilize a feature length of 512 for the codebook entries and set the codebook size to 256. The total decoding space for body gestures is represented as $\in \mathbb{R}^{T \times 256^3}$. The VQ-VAE is trained for 200 epochs, with a learning rate of 2.5$e$-4 for the initial 195 epochs, which is then decreased to 2.5$e$-4 for the last 5 epochs.

\paragraph{Global Motion Predictor} We train the Global Motion Predictor using an architecture that mirrors the CNN-based structure of our VQ-VAE's encoder and decoder The input consists of local motions and predicted foot contact labels $\in \mathbb{R}^{T \times 334}$, and it outputs global translations $\in \mathbb{R}^{T \times 3}$.

\section{Visualization Blender Add-on}
\label{secsup5}
For straightforward visualization of our BEAT2 dataset, we utilize the SMPL-X Blender add-on \cite{SMPL-X:2019}. As the latest SMPL-X add-on does not support the full range of facial expressions for SMPL-X, we extract 300 expression meshes from the original SMPL-X model and added them as individual blendshape targets into the SMPL-X model within the Blender add-on.

\section{Training time}
\label{secsup6}
We report the training time on a single L4, V100 and 4090 with a batch size (BS) of 64 for the best performance. 
\begin{table}[h]
\centering
\resizebox{0.90\linewidth}{!}{
\begin{tabular}{lcccccc}
~ ~ ~~                  & \multicolumn{2}{c}{1-speaker~}                              & \multicolumn{2}{c}{25-speaker~}         & \multicolumn{1}{l}{}     & \multicolumn{1}{l}{}    \\
\textbf{\textbf{EMAGE}} & \multicolumn{1}{l}{1-epoch} & \multicolumn{1}{l}{400-epoch} & 1-epoch & \multicolumn{1}{l}{100-epoch} & \multicolumn{1}{l}{Mem.} & \multicolumn{1}{l}{BS}  \\ 
\hline
L4 (24G)                & 239s                        & 26.5h                         & 3197s   & 89.6h                         & 20.1G                    & 64                      \\
V100 (32G)              & 155s                        & 17.2h                         & 2073s   & 58.1h                         & 20.1G                    & 64                      \\
4090 (24G)              & 72s                         & 8.0h                          & 963s    & 27.1h                         & 20.1G                    & 64                     
\end{tabular}}
\end{table}

Addtionally, pretraining of the 5 $\times$ VQVAEs for face, hands, upper body, lower body, and global motion would take 22.4 hours on 5 $\times$ 4090 GPUs.

\begin{table}[h]
\centering
\resizebox{0.90\linewidth}{!}{
\begin{tabular}{lcccccc}
\multicolumn{1}{c}{}        & \multicolumn{2}{c}{1-speaker} & \multicolumn{2}{c}{25-speaker} &       &     \\
\textbf{VQVAEs $\times$ 1 } & 1-epoch & 700-epoch           & 1-epoch & 100-epoch            & Mem.  & BS  \\ 
\hline
L4 (24G)                    & 200s    & 39.5h               & 2760s   & 74.4h                & 13.8G & 64  \\
V100 (32G)                  & 131s    & 25.5h               & 1727s   & 48.0h                & 13.8G & 64  \\
4090 (24G)                  & 61s     & 11.9h               & 802s    & 22.4h                & 13.8G & 64 
\end{tabular}}
\end{table}

\section{Importance of lower body motion}
\label{secsup7}
Lower body motion allows gestures \textbf{semantically aligned} with the content of the audio to achieve more vivid and impressive results, \textit{e.g.}, ``hiking in nature" with a walking gesture, ``playing football" with a kicking motion; see figure below. Compared with the upper body, it is more weakly related to the audio, but it still has connections in the above cases. 
\begin{figure}[h]
\centering
\includegraphics[width=1.0\columnwidth]{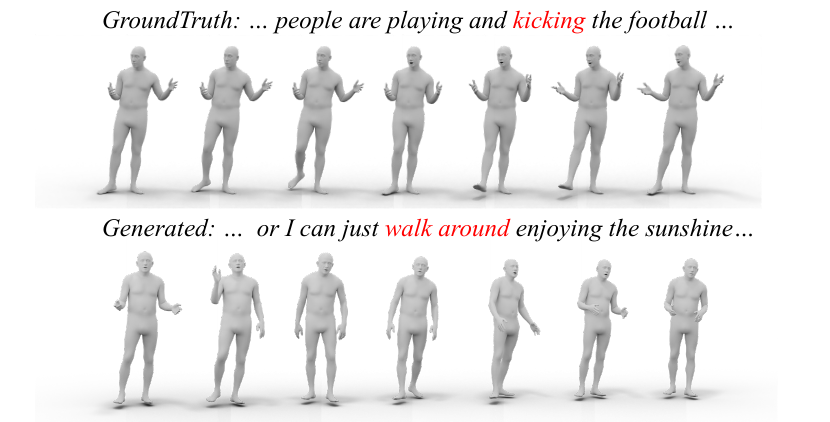}
\label{fig:fig6s}
\vspace{-0.2cm}
\end{figure}

In the implementation of EMAGE, we first obtain the latents of different body components with separate MLPs. Then, the lower body motion decoder \textbf{leverages all latents of ``audio", ``upper body", and ``hands"} for cross-attention based lower-body motion decoding. We have also observed that decoding directly from audio  increases diversity but reduces the coherence of the results on BEATv1.3. 

\begin{table}[h]
\centering
\resizebox{0.95\linewidth}{!}{
\begin{tabular}{lccccc}
                             & FGD $\downarrow$ & BC\textasciitilde{}$\uparrow$ & Diversity $\uparrow$ & MSE$\downarrow$ & LVD $\downarrow$  \\
\hline
audio only                       & 6.209                             & 6.683                                        & \textbf{13.714}                       & 1.183                                             & 8.788                                              \\
audio + upper + hands & \textbf{5.423}                    & \textbf{6.794}                               & 13.075                                & \textbf{1.180}                                    & \textbf{8.715}                                    
\end{tabular}}
\end{table}


\end{document}